\documentclass[letterpaper, 10 pt, conference]{ieeeconf}

\usepackage{graphicx}
\usepackage{balance}
\usepackage{comment}
\usepackage{cite}
\usepackage{amssymb}
\usepackage[tight,footnotesize]{subfigure}
\usepackage[active]{srcltx}
\usepackage{amsmath}
\usepackage{mathtools}

\graphicspath{{./figs/}}
\renewcommand{\baselinestretch}{0.95}
\usepackage{eurosym}
\usepackage[plain]{algorithm}
\usepackage{algorithmic}
\usepackage{multicol}
\usepackage{dsfont}
\usepackage{amsfonts}
\newcommand{\normvec}[1]{\left\lVert#1\right\rVert}
\usepackage{cases}
\usepackage{xcolor}
\IEEEoverridecommandlockouts
\begin{document}

\title{Integrated Ray-Tracing and Coverage Planning Control \\using Reinforcement Learning}

\author{Savvas~Papaioannou,~Panayiotis~Kolios,~Theocharis~Theocharides,\\~Christos~G.~Panayiotou~ and ~Marios~M.~Polycarpou
\thanks{The authors are with the KIOS Research and Innovation Centre of Excellence (KIOS CoE) and the Department of Electrical and Computer Engineering, University of Cyprus, Nicosia, 1678, Cyprus. {\tt\small \{papaioannou.savvas, pkolios, ttheocharides, christosp, mpolycar\}@ucy.ac.cy}}
}

\maketitle

\begin{abstract}
In this work we propose a coverage planning control approach which allows a mobile agent, equipped with a controllable sensor (i.e., a camera) with limited sensing domain (i.e., finite sensing range and angle of view), to cover the surface area of an object of interest. The proposed approach integrates ray-tracing into the coverage planning process, thus allowing the agent to identify which parts of the scene are visible at any point in time. The problem of integrated ray-tracing and coverage planning control is first formulated as a constrained optimal control problem (OCP), which aims at determining the agent's optimal control inputs over a finite planning horizon, that minimize the coverage time. Efficiently solving the resulting OCP is however very challenging due to non-convex and non-linear visibility constraints. To overcome this limitation, the problem is converted into a Markov decision process (MDP) which is then solved using reinforcement learning. In particular, we show that a controller which follows an optimal control law can be learned using off-policy temporal-difference control (i.e., Q-learning). Extensive numerical experiments demonstrate the effectiveness of the proposed approach for various configurations of the agent and the object of interest. 
\end{abstract}

\section{Introduction} \label{sec:Introduction}

The scientific advancements of the last decade in the areas of robotics, control, and machine learning have spurred an increased interest in the utilization of autonomous systems in various application domains including search-and-rescue \cite{Papaioannou2019,Liu2017,Papaioannou2020}, security \cite{PapaioannouJ1,Valianti2021,PapaioannouJ2}, and monitoring \cite{Corah2018,Zhou2017,Moon2021}. 

Coverage planning \cite{Galceran2013} is amongst the most important tasks found in many applications including infrastructure inspection, automated maintenance, area search, and surveillance, and thus plays a pivotal role in designing and executing automated missions using autonomous mobile agents. In coverage planning we are interested in finding a trajectory which allows an autonomous agent (e.g., a mobile robot) to observe (or cover) with its sensor every point/region within a specified area of interest. More specifically, during an automated coverage mission, the mobile agent must autonomously plan its coverage trajectory (e.g., determine its control inputs; possibly over a planning horizon), which allow the efficient coverage of the area of interest, while satisfying certain kinematic and sensing constraints.
As discussed in more detail in Sec. \ref{sec:Related_Work}, over the years, a plethora of coverage planning approaches have been proposed in the literature. However, there is still room for improvement until this technology reaches the required level of maturity. Specifically, the vast majority of coverage planning approaches mainly accounts for agents with uncontrollable and fixed sensors, and without considering the agent's kinematic and sensing constraints. This allows the coverage planning problem to be reduced to a conventional path planning problem which can further be transformed to some form of the traveling salesman problem i.e., seeking to find the shortest path that passes through a finite set of points.

Complementary to the existing literature, in this work we propose an integrated ray-tracing and coverage planning control approach where a mobile agent equipped with a controllable camera sensor with limited sensing domain (i.e., finite sensing range and angle of view), autonomously generates its trajectory such that the total surface area of a known object/area of interest is optimally covered. In the proposed approach we integrate ray-tracing into the planning process in order to determine the visible parts of the scene through the agent's camera sensor. 
The problem of integrated ray-tracing and coverage planning control is first formulated as constrained optimal control problem (OCP), which however is very challenging to be solved due to the existence of non-convex and non-linear constraints. For this reason, the problem is transformed into a finite Markov decision process (MDP) and is solved using reinforcement learning. Specifically, the contributions of this work are the following:

\begin{itemize}
    \item We propose a coverage planning control approach which allows an autonomous mobile agent to efficiently cover the total surface area of an object of interest by jointly optimizing its kinematic and camera control inputs. The proposed approach integrates ray-tracing into the coverage planning process in order to simulate the physical behavior of light-rays, thus enabling the agent to identify the visible parts of the scene through its camera field-of-view.    
    \item The problem of integrated ray-tracing and coverage control is initially formulated as a constrained optimal control problem and subsequently converted into a finite Markov decision process (MDP) which can be solved efficiently using reinforcement learning. Specifically, we show that off-policy temporal-difference control (i.e., Q-learning) can be utilized to learn a coverage controller which follows an optimal control law.
    \item Extensive numerical experiments demonstrate the effectiveness of the proposed approach for various configurations of the agent and the object of interest.

\end{itemize}

The rest of the paper is organized as follows. Section~\ref{sec:Related_Work} summarizes the related work on coverage planning control. Then, Section \ref{sec:system_model} develops the system model, Section \ref{sec:problem} formulates the problem tackled, and Section \ref{sec:approach} discusses the details of the proposed coverage planning approach. Finally, Section \ref{sec:Evaluation} evaluates the proposed approach and Section \ref{sec:conclusion} concludes the paper and discusses future work.

\section{Related Work}\label{sec:Related_Work}

Initial works on coverage planning control \cite{Choset1998,Acar2002,Acar2006,Huang2001,Mnnadiar2010} proceed by decomposing the environment into a finite number of disjoint cells, and then utilize a path-planning algorithm \cite{Nadhir2020} to find the best path that passes from every cell. The works in \cite{Li2005,Janchiv2013,Karapetyan2017,Cannata2011} investigate the coverage planning problem with multiple mobile agents which however exhibit fixed and uncontrollable sensor footprints. Different variations of the coverage planning problem are investigated in \cite{Schwager2009,Cortes2004,Pimenta2008} i.e., in \cite{Schwager2009,Cortes2004} the objective is to derive a decentralized, control law which guides a team of agents to an optimal configuration that maximizes coverage, and in \cite{Pimenta2008} the coverage planning problem is investigated with multiple heterogeneous agents.
The coverage planning problem is also investigated with the use of unmanned aerial vehicles (UAVs) \cite{Li2011,Xu2014,Liam2014,SavvasTAES}. Specifically, the authors in \cite{Li2011} propose an exact cellular decomposition method to plan the coverage path of UAVs equipped with fixed sensors in a polygon area, whereas in \cite{Xu2014,Liam2014} the problem is investigated under similar assumptions with fixed-wing UAVs. In \cite{SavvasTAES} the coverage planning problem is investigated in the presence of visibility constraints. The authors in \cite{Papaioannou2021a,Papaioannou2021b,Papaioannou2022} propose a search planning framework based on mixed integer quadratic programming (MIQP), which allows an autonomous UAV agent to cover in 3D specific objects of interest. Finally, in \cite{Theile2020,Sanna2021} learning based coverage planning techniques are investigated. Specifically, in \cite{Theile2020}, an end-to-end deep reinforcement learning coverage planning approach is proposed for an autonomous UAV agent with battery constraints, whereas in \cite{Sanna2021} the authors design a multi-agent coverage planning approach based on supervised imitation learning, for searching a finite number of cells within a bounded environment.




\section{System Model} \label{sec:system_model}

\subsection{Agent Kinematic Model} \label{ssec:kinematic_model}
We assume that an autonomous mobile agent, which in this work is represented by a point-mass object, evolves inside a bounded planar workspace $\mathcal{W} \subset \mathbb{R}^2$ according to a discrete-time kinematic model of the form $x_t = f(x_{t-1},u_{t-1})$, given by:

\begin{equation} \label{eq:kinematics}
x_t = x_{t-1} + d_R  \begin{bmatrix}
						\cos(\vartheta)\\
						\sin(\vartheta)
					\end{bmatrix}, ~ t \geq 1,~ x_0 = \bar{x}
\end{equation}

\noindent where  $x_t = [\text{x},\text{y}]^\top \in \mathbb{R}^2$ is the kinematic state of the agent at time $t$, which is given by the agent's position in 2D cartesian coordinates, and $\bar{x}$ is the agent's initial state. The agent's kinematic control input is given by $u_t = [d_R, \vartheta]$, where $d_R \in \mathbb{R}$ is the radial displacement, and $\vartheta \in [0, 2\pi)$ denotes the agent's heading. Finally, we assume that the agent's state is bounded i.e., $x_t \in \mathcal{X}, \forall t$ and the control input is constrained according the agent's kinematic capabilities i.e.,  $u_t \in \mathcal{U}, \forall t$.


\subsection{Agent Sensor Model} \label{ssec:sensing_model}

\begin{figure}
	\centering
	\includegraphics[scale=0.7]{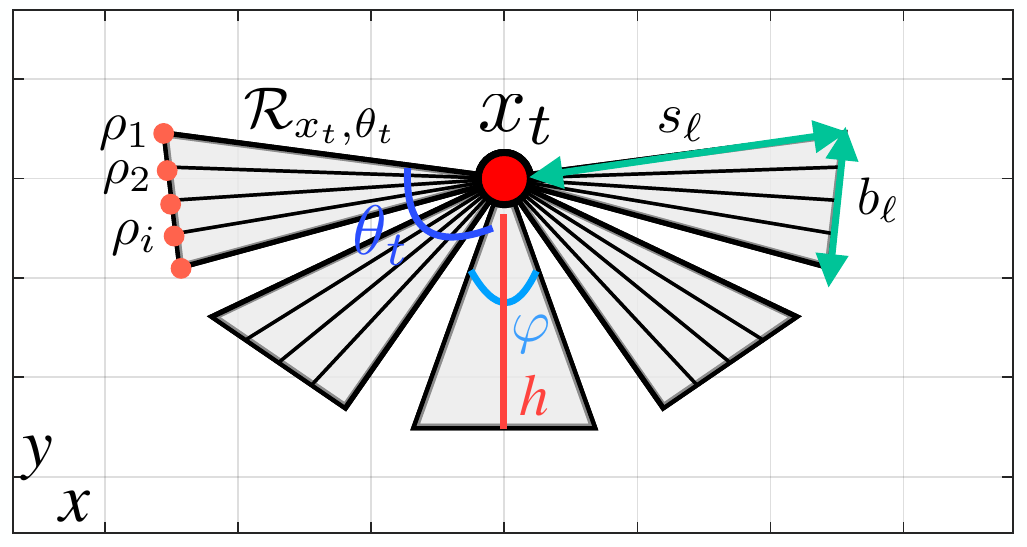}
	\caption{The figure illustrates the agent's camera FOV state for 5 rotation angles, where $x_t$ is the agent's kinematic state, $h$ is the camera sensing range, $\varphi$ is the FOV's angle of view, $s_\ell$ and $b_\ell$ are the FOV side length and base length respectively, and $\mathcal{R}_{x_t,\theta_t}$ denotes the set of camera rays at time $t$ when the camera FOV has been rotated by $\theta_t$.}	
	\label{fig:fig1}
	\vspace{-5mm}
\end{figure}

The autonomous mobile agent is equipped with a controllable camera sensor with limited sensing domain (i.e., finite sensing range and angle of view), which uses for observing the surrounding environment. The projected camera field-of-view (FOV) is modeled in this work as an isosceles triangle parameterized by its angle $\varphi$ at the apex and its height $h$, which are used to model the FOV's angle of view and sensing range respectively, as illustrated in Fig. \ref{fig:fig1}. The FOV's side length $(s_\ell)$ and base length $(b_\ell)$ are computed according to:
\begin{align}
    s_\ell &= h \times \text{cos}(\varphi/2)^{-1}, \quad b_\ell = 2 s_\ell \times \text{sin}(\varphi/2) 
\end{align}

\noindent and thus the matrix $\mathcal{F}_o$ which contains the vertices of a downward facing FOV with the apex at the origin is given by:
\begin{equation}
    \mathcal{F}_o =
    \begin{bmatrix}
       0 & -b_\ell/2 & b_\ell/2 \\
       0 & -h & -h \\
    \end{bmatrix}
\end{equation}
\noindent The agent's camera FOV is controllable i.e., it can be rotated around the agent's position $x_t$ by an angle $\theta_t \in \Theta \subseteq [0, 2\pi)$ at time $t$ according to: $\mathcal{F}_t = Q(\theta_t)\mathcal{F}_o + x_t$, where $\mathcal{F}_t$ is the rotated camera FOV state, and $Q(\theta_t)$ is a 2D rotation matrix which is given by:
\begin{equation} \label{eq:rotation_mat}
   Q(\theta_t) =
    \begin{bmatrix}
       \text{cos}(\theta_t) & \text{sin}(\theta_t) \\
       -\text{sin}(\theta_t) & \text{cos}(\theta_t) 
    \end{bmatrix}
\end{equation}

Moreover, we assume that at each time-step $t$ a finite set of light-rays (which are assumed to be straight), and which indicate the direction of the propagation of light, enter the camera's optical sensor and cause the matter inside the agent's FOV to be imaged. The set of light-rays entering the camera's optical sensor at time $t$ when the agent is at state $x_t$ and the camera FOV has been rotated by an angle $\theta_t$ is denoted as $\mathcal{R}_{x_t,\theta_t}=\{R_1,..,R_{|\mathcal{R}_{x_t,\theta_t}|}\}$, with $|.|$ to denote the set cardinality. The individual ray $R_i \in \mathcal{R}_{x_t,\theta_t}$ is given by the line-segment: $R_i = \{\rho_i + s(x_t-\rho_i) | s \in [0,1]\}$, where $x_t$ is the ray's end point on the optical sensor given by the agent's position at time $t$ and $\rho_i \in \mathbb{R}^2$ is a fixed point on the base of the camera's FOV and determines the ray's origin, as illustrated in Fig. \ref{fig:fig1}.

To summarize, given an initial kinematic state $x_0$, the agent's trajectory (i.e., a sequence of kinematic and FOV states) over a finite planning horizon of length $T$ time-steps can be optimized to meet certain objective by appropriately selecting the control inputs $\{\hat{u}_t | t=0,..,T-1\}$, where  $\hat{u}_t = [u_t,\theta_t]$.

\subsection{Object/Area of Interest Model}\label{ssec:object}
\begin{figure}
	\centering
	\includegraphics[scale=0.6]{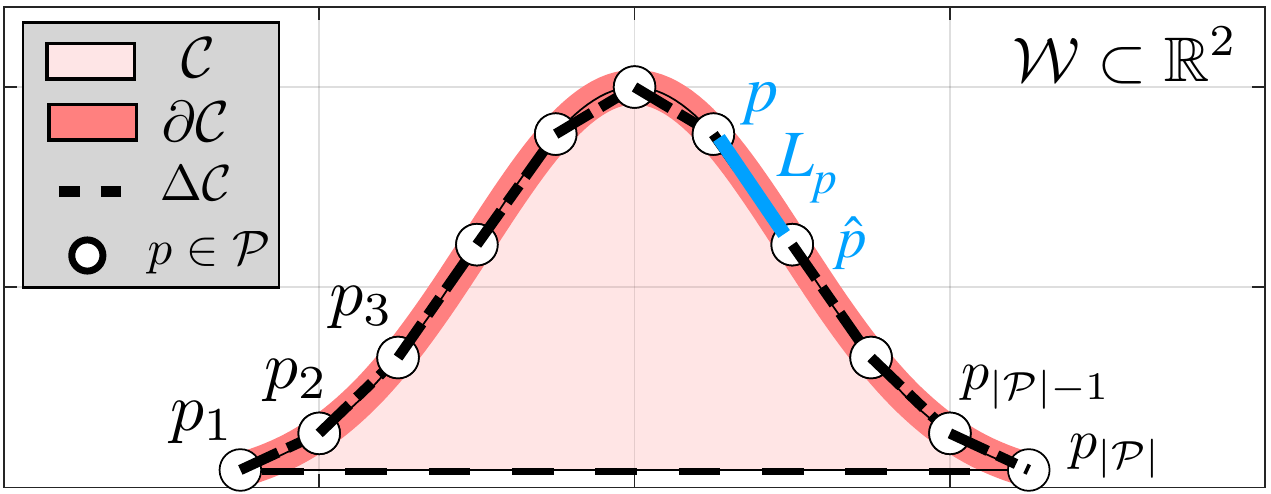}
	\caption{In the scenario above we are interested in finding the agent's optimal control inputs which result in the coverage of the object's ($\mathcal{C}$) boundary $\partial \mathcal{C}$. $\Delta \mathcal{C}$ is the piece-wise linear approximation of $\partial \mathcal{C}$, which is composed of a finite number of line segments $L_{p}$ connecting the points $p \in \mathcal{P}$.  }	
	\label{fig:fig2}
	\vspace{-5mm}
\end{figure}

The objective of our autonomous mobile agent is to optimally decide its kinematic and camera FOV inputs $\{\hat{u}_t | t=0,..,T-1\}$ over a finite planning horizon such that the total surface area of a bounded convex object or area of interest $\mathcal{C} \subset \mathcal{W}$ is covered by its sensor's FOV. More precisely, we are interested in the coverage of the object's boundary $\partial\mathcal{C}$, as illustrated in Fig. \ref{fig:fig2}, since we are operating in a 2D environment. However, the problem formulation and the proposed approach can be generalized in 3D space as well.

The boundary $\partial\mathcal{C}$ of the object of interest $\mathcal{C}$ is approximated in this work by a piece-wise linear model denoted as $\Delta \mathcal{C}$, which is obtained by connecting together a number of points $p \in \partial \mathcal{C}$ (uniformly) sampled from the boundary. The number of points sampled is chosen according to the mission requirements (e.g., required coverage precision). Specifically, we denote the finite set of points sampled from the object's boundary as $\mathcal{P} = \{p_1,..,p_{|\mathcal{P}|}\} \subset \partial \mathcal{C}$. Subsequently, the piece-wise linear approximation $\Delta \mathcal{C}$ of the boundary $\partial \mathcal{C}$ is composed of line segments: $L_{p}= \{p+ r (\hat{p}-p)| r \in [0,1], \hat{p} \ne p \in \mathcal{P}\}$, which have been generated by connecting the points $(p,\hat{p})_{p \ne \hat{p}} \in \mathcal{P}$, such that the resulting line segments $L_{p} \forall p$ belong to the boundary of the convex hull of $\mathcal{P}$, as shown in Fig. \ref{fig:fig2}. As we discuss in more detail in Sec. \ref{sec:problem}, the agent's objective now becomes the coverage of all points $p \in \mathcal{P}$ on the object's boundary.

%

\section{Problem Formulation}\label{sec:problem}

The problem of integrated ray-tracing and coverage planning control tackled in this work can now be formulated as an optimal control problem (OCP) shown in Problem (P1).

\begin{algorithm}
\begin{subequations}
\begin{align}
&\hspace*{-1mm}\textbf{Problem (P1):}~\texttt{Optimal Controller} &  \nonumber\\
& \hspace*{-1mm}~~~~\underset{\mathbf{\hat{U}}_T}{\arg \max} ~\mathcal{J}_\text{coverage} &\label{eq:objective_P1} \\
&\hspace*{-1mm}\textbf{subject to: $t=[1,..,T]$} ~  &\nonumber\\
&\hspace*{-1mm}  x_{t} = f(x_{t-1},u_{t-1}) & \forall t \label{eq:P1_1}\\
&\hspace*{-1mm} x_0 = \bar{x} & \label{eq:P1_2}\\
&\hspace*{-1mm} \exists ~\tau \in [1,..,T] : \left( p \in \mathcal{F}_\tau \right) \wedge \mathds{1}_{G_\tau}(p) & \hspace*{-5mm} \forall p \in \mathcal{P}\label{eq:P1_3}\\
&\hspace*{-1mm} x_0, x_{t} \notin \mathcal{C} &\forall t \label{eq:P1_4}\\
&\hspace*{-1mm} x_0, x_{t} \in \mathcal{X} & \forall t\label{eq:P1_5}\\
&\hspace*{-1mm} \hat{u}_{t} =[u_{t},\theta_{t}] \in \mathcal{U} \times \Theta  & \forall t\label{eq:P1_6}
\end{align}
\end{subequations}
\vspace{-5mm}
\end{algorithm}

\noindent The objective is to find the optimal control inputs $\mathbf{\hat{U}}_T=\{\hat{u}_0,..,\hat{u}_{T-1}\}$, subject to the constraints in Eqn. \eqref{eq:P1_1}-\eqref{eq:P1_6}, over a finite planning horizon of length $T$ time-steps, which will result in the maximization of the coverage performance objective $\mathcal{J}_\text{coverage}$ i.e., Eqn. \eqref{eq:objective_P1}. 
As we have already discussed in Sec. \ref{ssec:kinematic_model}, the constraints in Eqn. \eqref{eq:P1_1}-\eqref{eq:P1_2} are due to the agent's kinematic model. Then, the constraint in Eqn. \eqref{eq:P1_3} implements the coverage functionality. That is, for every point $p \in \mathcal{P}$ on the object's boundary $\partial \mathcal{C}$, it seeks to find a time-step $\tau \in [1,..,T]$ inside the planning horizon, where the applied control inputs $\{\hat{u}_0,..,\hat{u}_\tau\}$ generate a trajectory (i.e., a sequence of kinematic and FOV states) which result in the coverage of point $p$ at time $\tau$ (i.e., $p \in \mathcal{F}_\tau$). Subsequently, the second part of the constraint in Eqn. \eqref{eq:P1_3} requires that at time-step $\tau$, at which point $p$ resides inside the agent's FOV, the same point $p$ must also be visible i.e., overall Eqn. \eqref{eq:P1_3} requires that point $p$ must reside within the visible FOV of the agent. This constraint is required to account for the fact that the agent's FOV at some time-step $\tau \in [1,..,T]$ might be blocked by obstacles, prohibiting the light-rays from reaching the camera's optical sensor and therefore rendering certain parts of the scene unobservable. In essence the indicator function $\mathds{1}_{G_\tau}(p)$, utilizes the function $G_\tau(p)$ in order to trace light-rays back to their source, and thus determining the existence of a light-ray which traces-back to point $p$, consequently determining $p$'s visibility on the agent's camera. Thus $\mathds{1}_{G_\tau}(p)$ is defined in this work as:
\begin{equation}
 \mathds{1}_{G_\tau}(p) =
  \begin{cases}
    1,       & \quad \text{if } G_\tau(p) \ne \emptyset,\\
    0,      & \quad \text{o.w. }
  \end{cases}
\end{equation}

Next, the constraint Eqn. \eqref{eq:P1_4} implements a collision avoidance constraint with the object of interest $\mathcal{C}$ i.e., the kinematic state $x_t$ of the agent must always reside outside of the object's boundary $\partial \mathcal{C}$. Finally, the constraints in Eqn. \eqref{eq:P1_5} and \eqref{eq:P1_6} restrict the agent's state and control inputs within the desired operating bounds. Next, we provide a more detailed discussion on how the formulation discussed above can be tackled.

\subsection{Visible FOV Determination}
In this section we describe in more detail how the visible parts of the agent't FOV can be determined using ray-tracing. Essentially, we are interested in determining whether the propagation of light-rays have been blocked by obstacles and thus rendering parts of the scene unobservable. This procedure is also useful in determining which of the points $p \in \mathcal{P}$ on the object's boundary belong to the foreground and which points belong to the background (and thus are unobservable from a specific agent state). 

Recall that the set of all light-rays entering the camera's optical sensor at time $t$ is given by $\mathcal{R}_{x_t,\theta_t}$, where $x_t$ and $\theta_t$ is the agent's kinematic state and FOV rotation angle at time $t$ respectively. Additionally, as we have already mentioned in Sec. \ref{ssec:object}, the boundary $\partial \mathcal{C}$ of the object of interest has been piece-wise linearly approximated  with a finite number of line segments $\Delta \mathcal{C} = \{L_{p}| p \in \mathcal{P}\}$. We can now determine whether point $p \in \mathcal{P}$ belongs to the visible FOV at time $t$, with respect to the light-rays in $\mathcal{R}_{x_t,\theta_t}$ by finding at least one light-ray $\hat{R} \in \mathcal{R}_{x_t,\theta_t}$ which traces back to point $p$ as follows:
\begin{equation}
    G_t(p) = \{\hat{R} \in \mathcal{R}_{x_t,\theta_t}: \hat{R} \otimes \Delta \mathcal{C} = \hat{L}_{p}\}
\end{equation}

\begin{figure}
	\centering
	\includegraphics[width=\columnwidth]{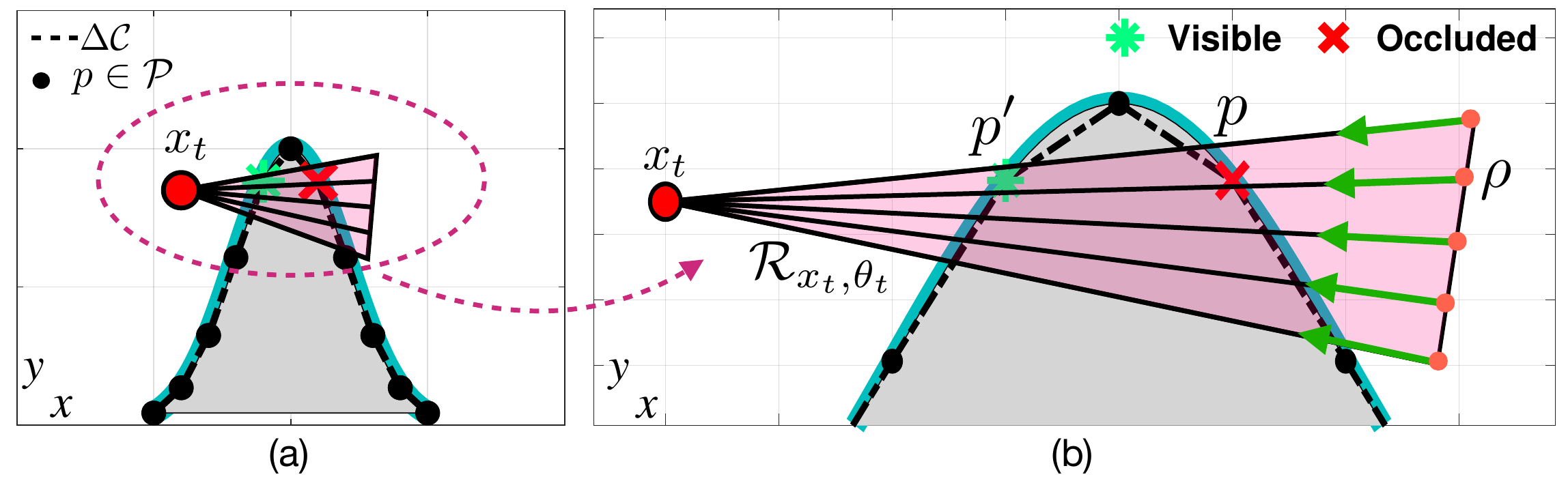}
	\caption{The figure illustrates the ray-tracing procedure utilized in the proposed coverage planning approach for identifying the visible parts of the scene through the agent's camera. In the above figure point $p^\prime$ is visible, whereas point $p$ is occluded. The propagation of rays are from $\rho \rightarrow x_t$.}	
	\label{fig:fig3}
	\vspace{-5mm}
\end{figure}

\noindent where the operation $R \otimes \Delta \mathcal{C}$ defines the intersection of the light-ray $R$, with all line segments  $L_p \in \Delta \mathcal{C}$, and returns the line segment $\hat{L}_p$ (if exists) which is the closest to the camera's optical sensor where the light-ray $R$ converges (i.e., see Fig.\ref{fig:fig3}). Stated differently, $\hat{L}_p$ is the line-segment which light-ray $R$ intersects last. Since, light-rays propagate towards the camera's optical sensor where they converge, the line-segment encountered the last by a light-ray is the line-segment $\hat{L}_p$ which belongs to the visible FOV. For instance this is the line segment on the boundary of the object of interest which can be imaged through the agent's camera. On the other hand the line-segment which first intersects with the light-ray (assuming the light-ray intersects with more than one line-segments) is not visible e.g., this line-segment might belong to the invisible side of the object as depicted in Fig. \ref{fig:fig3}. Consequently, the point $p$ on the visible line-segment $\hat{L}_p$ is also visible.

Let now the camera-ray $R$ given by $R = \{\rho + s(x_t-\rho) | s \in [0,1]\}$, to have  $x$ and $y$ cartesian coordinates given by $\rho(x) + s[x_t(x) - \rho(x)]$ and $\rho(y) + s[x_t(y) - \rho(y)]$ respectively. Also, let the $x$ and $y$ cartesian coordinates of the a line-segment $L_{p} = \{p + r (\hat{p}-p)| r \in [0,1], \hat{p} \ne p\}$ to be $p(x) + r[\hat{p}(x) - p(x)]$ and $p(y) + r[\hat{p}(y) - p(y)]$ respectively. The intersection $R \otimes L_{p}$ of the light-ray $R$ with the line segment $L_p$  can be computed by solving the following system of linear equations for the two unknowns i.e., $(s,r)$:
\begin{equation}\label{eq:linearSystem}
    \begin{bmatrix}
    x_t(x)-\rho(x) &  p(x)-\hat{p}(x) \\
    x_t(y)-\rho(y) &  p(y)-\hat{p}(y)
   \end{bmatrix}
   \begin{bmatrix}
    s\\
    r
   \end{bmatrix} = 
   \begin{bmatrix}
    p(x)-\rho(x)\\
    p(y)-\rho(y)
   \end{bmatrix}
\end{equation}

\noindent An intersection between $R$ and $L_p$ exists only when the solution $(s,r) \in [0,1]$. It should be noted here that the event of no intersection  implies no visibility. Summarizing, the agent with kinematic state $x_t$, and FOV state $\mathcal{F}_t$ (parameterized by angle $\theta_t$) observes and covers point $p \in \mathcal{P}$ at time $t$ when: a) $p$ resides inside the agent's camera FOV $p \in \mathcal{F}_t$ and b) there exists a light-ray $R \in \mathcal{R}_{x_t,\theta_t}$ which traces back to point $p$ i.e., the last intersection of $R$ is with the line-segment $L_{p}$ which contains point $p$. This is exactly what the constraint in Eqn. \eqref{eq:P1_3} describes.

\subsection{Collision Avoidance Constraints}\label{ssec:collisionAvoidance}
Collision avoidance between the agent and the object of interest is implemented in this work by making sure that the agent's kinematic state $x_t, \forall t$ always resides outside the convex hull of the object's $\mathcal{C}$ polygonal region approximation defined by its boundary $\Delta \mathcal{C}$ as:

\begin{equation} \label{eq:colision1}
    \exists p \in \mathcal{P}: \alpha^\top_{p} x_t >  \beta_p
\end{equation}

\noindent where $(a_p,\beta_p)$ are the coefficients (i.e., $a_p$ is the outward normal vector on $L_p$, and  $\beta_p$ is a constant) of the line equation which contains the line segment $L_p, p \in \mathcal{P}$. Since the area inside the object of interest is modeled as a convex polygonal region with boundary $\Delta \mathcal{C}$ defined by $|\mathcal{P}|$ linear equations containing the line segments $L_p$, a collision occurs at time $t$ when $\alpha^\top_{p} x_t \leq  \beta_p, \forall p$. Thus, a collision is avoided if there exists $p \in \mathcal{P}$ such that $\alpha^\top_{p} x_t >  \beta_p$. This can be implemented with the following set of constraints:
\begin{algorithm}
\vspace{-5mm}
\begin{subequations}
\begin{align}
& \alpha^\top_{p} x_t > \beta_{p} - M b^p_{t} ,~\forall t,p \label{eq:O_1}\\
& \sum_{p=1}^{|\mathcal{P}|} b^p_{t} \le  (|\mathcal{P}|-1), ~ \forall t, \label{eq:O_2}
\end{align}
\end{subequations}
\vspace{-5mm}
\end{algorithm}

\noindent where the variable $b^p_{t} \in \{0,1\}$ is a binary decision variable, and $M$ is a large positive constant. Essentially, Eqn. \eqref{eq:O_1} first checks if for some time $t \in [1,..,T]$, the $p_\text{th}$ equation is true i.e., $\alpha^\top_{p} x_t >  \beta_p$ and if so, it sets the binary decision variable $b^p_{t}=0$. On the other hand, if $\alpha^\top_{p} x_t \leq  \beta_p$, then $b^p_{t}$ is activated in order to make sure that Eqn. \eqref{eq:O_1} is satisfied. Subsequently, Eqn. \eqref{eq:O_2} requires that the number of times $b^p_{t}$ is activated at each time $t$ is less or equal to $(|\mathcal{P}|-1)$ which signifies that the agent's kinematic state resides outside the area enclosed by the object of interest i.e., $x_t \notin \mathcal{C}, \forall t$.

\subsection{Coverage Performance Objective}
Finally the coverage performance objective i.e., $\mathcal{J}_\text{coverage}$ is defined in this work as:
\begin{equation}
    \mathcal{J}_\text{coverage}= \sum_{t=1}^T J(\mathcal{F}_t,G_t)\sigma(t)
\end{equation}
\noindent where $\mathcal{F}_t$ is the agent's FOV state at time $t$ (associated with the agent's kinematic state $x_t$), $G_t$ is the ray-tracing function, and $J(\mathcal{F}_t,G_t)$ is defined as:

\begin{equation}\label{eq:stageCost}
    J(\mathcal{F}_t,G_t) =\sum_{p \in \mathcal{P}} \ell(\mathcal{F}_t,G_t,p)
\end{equation}

\noindent The function $\ell(\mathcal{F}_\tau,G_\tau,p) \in \{0, 1\}$ is further given by:

\begin{equation}
 \ell(\mathcal{F}_t,G_t,p) =
  \begin{cases}
    1, &  \exists \tau \leq T : \left( p \in \mathcal{F}_\tau \right) \wedge \mathds{1}_{G_\tau}(p),\\
    0,      &  \text{o.w. }
  \end{cases}
\end{equation}

\noindent In essence $\mathcal{J}_\text{coverage}$ is maximized when all points $p \in \mathcal{P}$ become visible at some point in time inside the planning horizon and they are covered by the agent's camera. Finally, the time-dependent term $\sigma(t)$ is used here to penalize points that are covered later in the planning horizon e.g., $\sigma(t) = (T-t)T^{-1}$, thus encouraging the agent to finish the coverage mission as soon as possible.

\section{RL-based Integrated Ray-tracing and Coverage Control}\label{sec:approach}

The optimal control problem i.e., Problem (P1), presented in the previous section is quite challenging to be solved efficiently in real-time, mainly due to the non-linear and non-convex constraints introduced when someone attempts to solve Eqn. \eqref{eq:linearSystem}, which is required by the constraint in Eqn. \eqref{eq:P1_3}. It should also be noted here that the computational complexity of the OCP in (P1) grows at best linearly with the size of the planning horizon; which can introduce additional challenges for large problems. Finally, the size of the planning horizon depends on the coverage scenario (e.g., the size of the object of interest, the initial location of the agent, etc.), and thus cannot be easily determined and tuned a-priori. If the planning horizon is chosen too short the problem becomes infeasible and the coverage mission fails. On the other hand, if the size of the planning horizon is overestimated, unnecessary computational complexity will be introduced. In order to tackle some of the challenges mentioned above, the optimal control problem in (P1) is approximated and transformed into a finite Markov decision process (MDP) for which an optimal control policy can be learned from experience using reinforcement learning (RL).

\subsection{Problem re-formulation using MDP}

The problem of integrated ray-tracing and coverage planning control is re-formulated in this section as a Markov decision process, which is then solved using reinforcement learning. 
A finite Markov Decision Process (MDP) \cite{MDP} is a discrete-time stochastic control process which can be defined with a set of components i.e., a tuple $\langle \mathcal{S}, \mathcal{A}, \mathcal{T}, R_w \rangle$; where $\mathcal{S}$ represents the state-space of the system (i.e., a finite set describing all possible states of the system), $\mathcal{A}$ is the action space (i.e., the finite set of actions available to the  agent), $\mathcal{T}$ is a discrete-time state transition function i.e., $\mathcal{T}:\mathcal{S} \times \mathcal{A}\mapsto \mathcal{S}$, which describes the evolution of the system due to actions, and finally $R_w:\mathcal{S} \times \mathcal{A}\mapsto \mathbb{R} $ is the reward function which returns the immediate reward when action $a$ is executed at state $s$ and the system transitions to some new state $s^\prime$.
Reinforcement learning \cite{Sutton1992} can be used to solve an MDP by finding a control policy (i.e., a mapping from states to actions) which maximizes the cumulative reward. In order to do that, RL methods track the interactions of the agent with the environment, and learn to optimize the control policy using reinforcement, i.e., a feedback in the form of reward or punishment. RL methods are particularly useful for solving a) large MDP problems, which are otherwise intractable with exact dynamic programming methods, and b) problems where the exact mathematical model of the MDP is not known. We can now go ahead and discuss how we have transformed our problem to an MDP. More specifically, the MDP components i.e., $\langle \mathcal{S}, \mathcal{A}, \mathcal{T}, R_w \rangle$ are constructed for our problem as follows:

\subsubsection{State-space $\mathcal{S}$} In order to construct the system's state-space we first decompose the environment $\mathcal{W}$ into a finite number of non-overlapping cells $\{c_1,..,c_{|\hat{\mathcal{W}}|}\}$, essentially creating a uniform grid $\hat{\mathcal{W}} = \bigcup_{i=1}^{|\hat{\mathcal{W}}|} c_i$, which within the agent can evolve by moving from one cell to another. 
Let $\bar{p}$ to denote the center of mass (i.e., centroid) of the object of interest $\mathcal{C}$, given by $\bar{p} = |\mathcal{P}|^{-1} \sum_{i=1}^{|\mathcal{P}|} p_i$. Let $\hat{d}_t= q(\normvec{x_t-\bar{p}}_2)$ to denote the quantized Euclidian distance between the agent's kinematic state and the object's center of mass $\bar{p}$, where the quantizer function $q()$ maps the distance values into the finite set $\hat{D}$. Subsequently the MDP state-space can now be defined as:
\begin{equation}
    \mathcal{S} = \hat{\mathcal{W}} \times \hat{P} \times \hat{D}
\end{equation}

\noindent where $\hat{P} = [0,..,|\mathcal{P}|]$, with $\mathcal{P}$ denoting the set of points sampled from the object's boundary. Therefore, a particular state $s_t = [s_a,s_b,s_c] \in \mathcal{S}$ at time $t$ describes the agent's kinematic state i.e., position $s_a \in \tilde{\mathcal{W}}$, the total number of points $s_b \in \hat{P}$ that are observed and covered for the first time from state $s_a$, and the agent's distance $s_c \in \hat{D}$ from the object's centroid.

\subsubsection{Action-space $\mathcal{A}$} The agent's kinematic model and control inputs are discretized in space according to:

\begin{equation} \label{eq:kinematics2}
\hat{x}_t = \hat{x}_{t-1} + l_R\Delta_R \begin{bmatrix}
						\cos(l_\vartheta \Delta_\vartheta)\\
						\sin(l_\vartheta \Delta_\vartheta)
					\end{bmatrix},  
					\begin{array}{l} 
						l_\vartheta = 0,...,N_\vartheta\\ 
						l_R = 0,...,N_R
				    \end{array} 
\end{equation}

\noindent where the agent's kinematic control input has now become $u^d_t = [l_R\Delta_R, l_\vartheta \Delta_\vartheta]$, where $\Delta_R$ is the radial step size, $\Delta_\vartheta=2\pi/N_\vartheta$, and the parameters $(N_\vartheta,N_R)$ are used to specify the total number of admissible control actions, denoted as $\mathcal{U}^d$. Specifically, the parameters $(\Delta_R,N_\vartheta,N_R)$ are used to ultimately determine the evolution of the agent's kinematic state $\hat{x}_t$ inside the discretized representation of the world. The camera's FOV rotation signal  is also discretized to take its values from within a finite set of rotation angles $\theta^d_t \in \Theta^d$. Consequently, the camera FOV state $\mathcal{F}_t$ at time $t$, also takes its values from within a finite set of possible FOV configurations. The agent's control signal, referred to as the agent's action hereafter, is denoted as $a_t = [u^d_t, \theta^d_t] \in \mathcal{U}^d \times \Theta^d$

\subsubsection{Reward $R_w$} Finally, we have designed a reward function $R_w(s_t,a_t)$ which closely resembles the constraints and objectives of the optimal control problem presented in Sec. \ref{sec:problem}, and which is given by:
\begin{equation}\label{eq:reward}
    R_w(s_t,a_t) = -w_1 - w_2\mathds{1}_{s_{t+1} \in \mathcal{C}} + w_3 J(s_{t+1})
\end{equation}

\noindent where the notation $(s_t,a_t,s_{t+1})$ refers to current state of the agent $s_t$, the action  $a_t$ applied at state $s_t$, and the transition of the agent to the new state $s_{t+1}$ in the next time-step. The parameters $w_1, w_2$ and $w_3$ are tuning weights which control the system's reinforcement behavior. More specifically, $w_1$ is a punishment term which used in Eqn. \eqref{eq:reward} in order to penalize the agent for every time-step that passes and the object of interest remains not fully covered. In other words this term encourages the agent to finish the coverage mission as soon as possible. Next, the indicator function $\mathds{1}_{s_{t+1} \in \mathcal{C}}$, which is weighted by $w_2$, uses the procedure described in Sec. \ref{ssec:collisionAvoidance} to identify whether the state-action pair $(s_t,a_t)$ results in a collision between the agent and the object of interest at the next time-step, and if so the agent is penalized with $-w_2$. Finally, the last term rewards the agent for its successful coverage efforts. Specifically, in the last term we make use of the function $J()$ described in Eqn. \eqref{eq:stageCost} to count the number of points which have been covered for the first time when the agent  transitions to the new state $s_{t+1}$; and we generate a reward signal weighted by $w_3$ as shown. 

\subsection{Off-policy Temporal-Difference Control}

Based on the MDP formulation discussed above the problem of integrated ray-tracing and coverage planning control, shown in (P1), can now be written as:
\begin{equation}\label{eq:q1}
 Q_\pi(s,a) = \mathbb{E}_\pi \left[ \sum_{t=0}^{\infty} \gamma^t R_w(s_t,a_t) | s_0=s, a_0=a \right]
\end{equation}

\noindent where $\gamma \in [0,1]$ is a discount factor which determines the importance of future rewards. The state-action value function $Q_\pi(s,a)$ measures the expected infinite horizon discounted cumulative reward when the agent starts from state $s$, takes action $a$ and then follows the control policy $\pi$. Subsequently, the optimal state-action value function is obtained as $Q_\star(s,a) = \max_\pi Q_\pi(s,a)$ by following an optimal control policy $\pi_\star$ which always chooses the optimal action in every state. The problem re-formulation discussed so far is the MDP equivalent of the maximization of the objective in Eqn. \eqref{eq:objective_P1} under the constraints \eqref{eq:P1_1}-\eqref{eq:P1_6}, i.e., Problem (P1). In this work, we use off-policy temporal-difference control or otherwise known as Q-learning \cite{Sutton1999}, in order to learn an optimal control policy $\pi_\star$ from experience. In particular, Q-learning is an iterative bootstrapping process which uses temporal-differencing in order to learn the state-action value function as: $Q(s_t,a_t)=$
\begin{equation}\label{eq:q2}
    Q(s_t,a_t) + \alpha \left[R_w(s_t,a_t) + \gamma \max_a Q(s_{t+1},a)-Q(s_t,a_t)\right] \notag
\end{equation}

\noindent where $\alpha$ is the learning rate. Being an off-policy method, Q-learning is able to learn the state-action value function $Q(s,a)$, which directly approximates $Q_\star(s,a)$, i.e.,  the optimal state-action value function, independent of the policy being followed. When the optimal state-action value function $Q_\star(s,a)$ is obtained, the optimal policy $\pi_\star$ can be derived by taking at each time-step the actions that maximize $Q_\star(s,a)$. Finally, Q-learning guarantees that an optimal policy can be found for any finite MDP given enough exploration time and a partly-random policy \cite{Watkins1992}.

\section{Evaluation} \label{sec:Evaluation}
\subsection{Simulation Setup}

The simulation setup used for the evaluation of the proposed approach is as follows: 
The surveillance area $\mathcal{W}$ of size $20$m by $20$m is converted into a uniform grid $\hat{\mathcal{W}}$ of size 10 by 10. At any time-step $t$, the quantized distance $\hat{d}_t$ between the agent and the centroid of the object of interest takes values from within the set $\hat{D}=\{0,0.5,1,1.5,...,\lceil \sqrt{2}\times20 \rceil\}$m. The agent kinematics follow Eqn. \eqref{eq:kinematics2} with $\Delta_R=2$m, $N_\vartheta=8$, and $N_R=1$. The agent's camera angle of view angle $\varphi$ is set to 40deg, and the sensing range $h=10$m. In total we consider 5 camera rotation angles $\theta^d$ which take values in the set $\Theta^d=\{-85, -42.5, 0, 42.5, 85\}$deg as shown in Fig. \ref{fig:fig1}. At each time-step $t$ and for each camera FOV configuration 5 light-rays enter the optical sensor i.e., $|\mathcal{R}_{\hat{x}_t,\theta^d_t}|=5$. The region/object of interest $\mathcal{C}$ is represented by a bell-shaped curve (as illustrated in Fig. \ref{fig:res2}), given by $f(x) = a\times\text{exp}\left(\frac{-(x-b)^2}{2c^2} \right)$ where $a, b$ and $c$ are free parameters. The objective is to to cover a total of 11 points $\mathcal{P}=\{p_1,..,p_{11}\}$ uniformly sampled from the object's boundary $\partial \mathcal{C}$. The punishment and reward parameters $w_1, w_2$ and $w_3$ in Eqn. \eqref{eq:reward} are set as 1, 100, and 2 respectively. The Q-learning parameters $\alpha$ and $\gamma$ (i.e., learning-rate and discount factor) are set to 0.1 and 0.8 respectively. We use a decreasing $\epsilon$-greedy exploration strategy with an initial $\epsilon=0.9$ which decays in every time-step as $\epsilon_{t+1}=0.9999\epsilon_{t}$. Finally,  for each episode the agent is randomly initialized inside $\hat{\mathcal{W}}$, and interacts with the environment until the terminal condition is met i.e., all points $|\mathcal{P}|$ have been covered or the time limit of 100 steps has been reached.

\subsection{Performance Evaluation}

\begin{figure}
	\centering
	\includegraphics[width=\columnwidth]{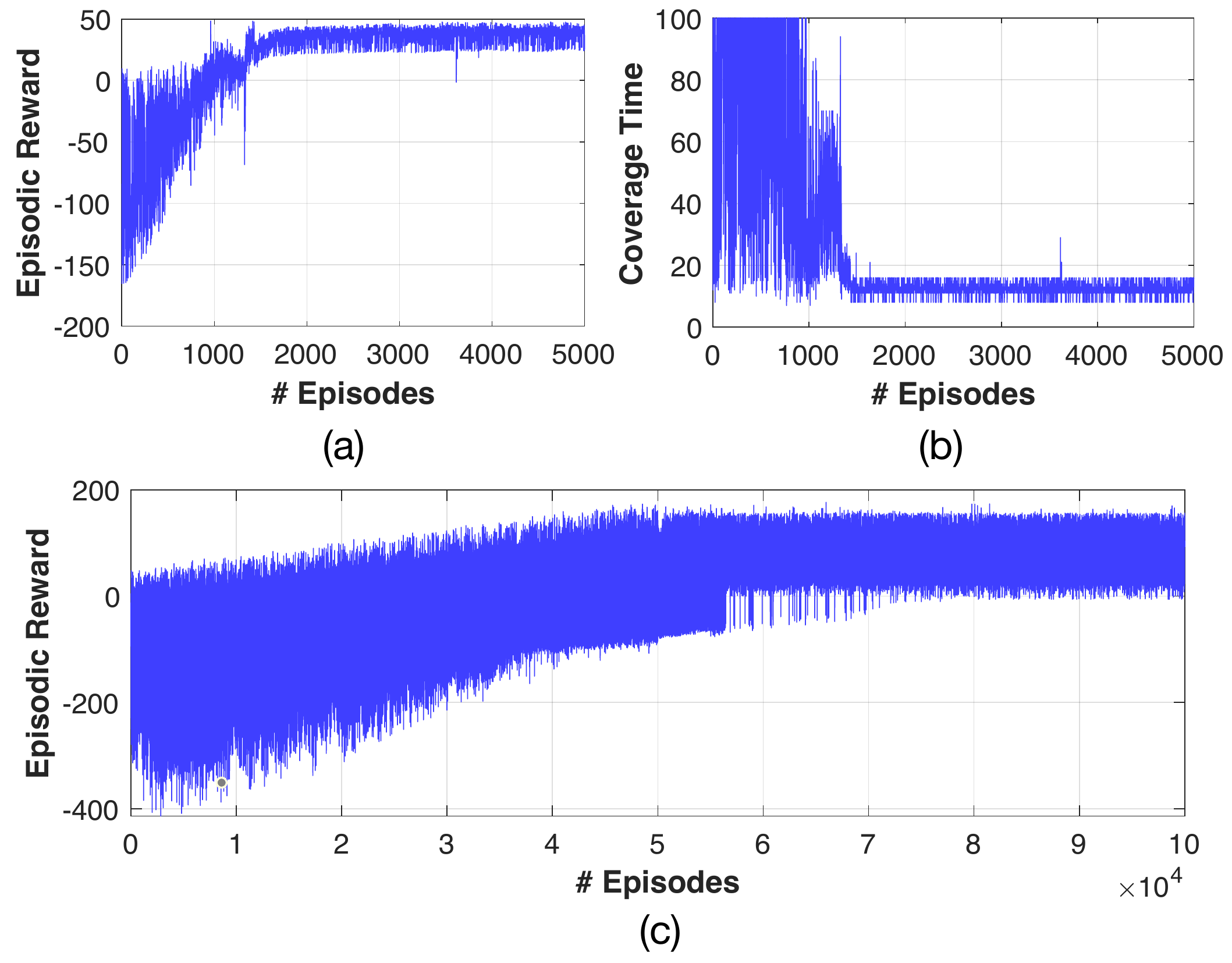}
	\caption{The figure shows the cumulative reward obtained in each episode during training. (a) The episodic reward obtained during training with a single configuration of the object of interest, and (b) the corresponding coverage time. (c) The evolution of the episodic reward during the training with multiple random configurations of the object of interest.}	
	\label{fig:res1}
	\vspace{-0mm}
\end{figure}

\begin{figure}
	\centering
	\includegraphics[scale=0.60]{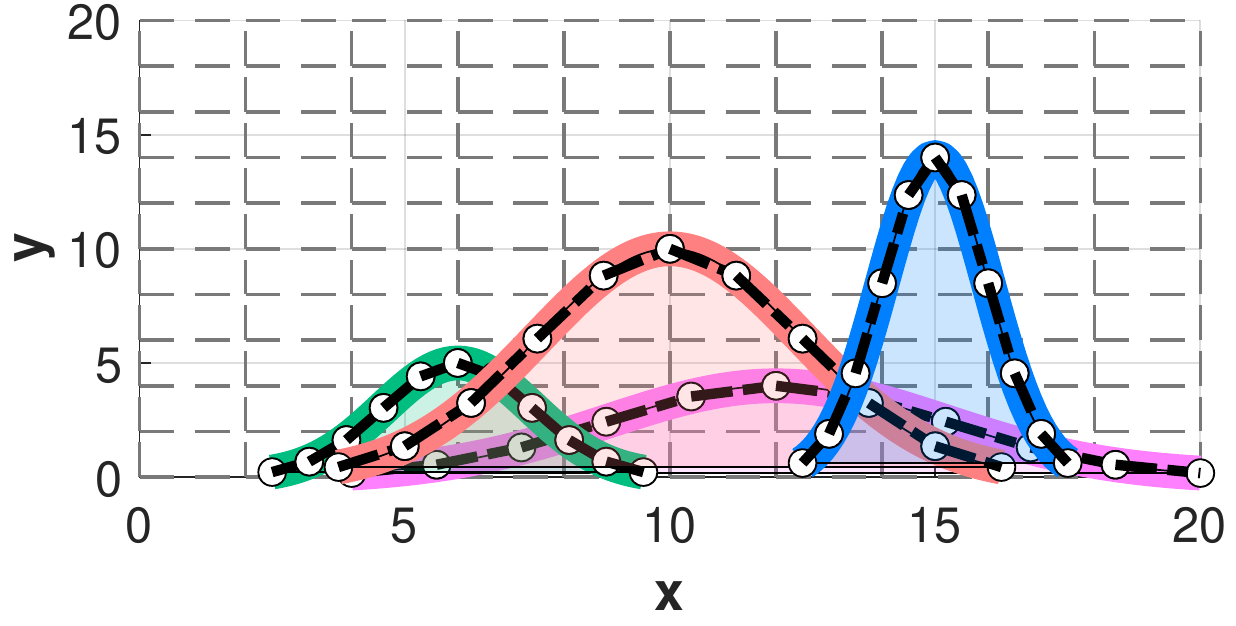}
	\caption{The figure shows an illustrative example of four different random realizations of the object of interest that have been sampled during training.}	
	\label{fig:res2}
	\vspace{-6mm}
\end{figure}

\begin{figure*}
	\centering
	\includegraphics[width=\textwidth]{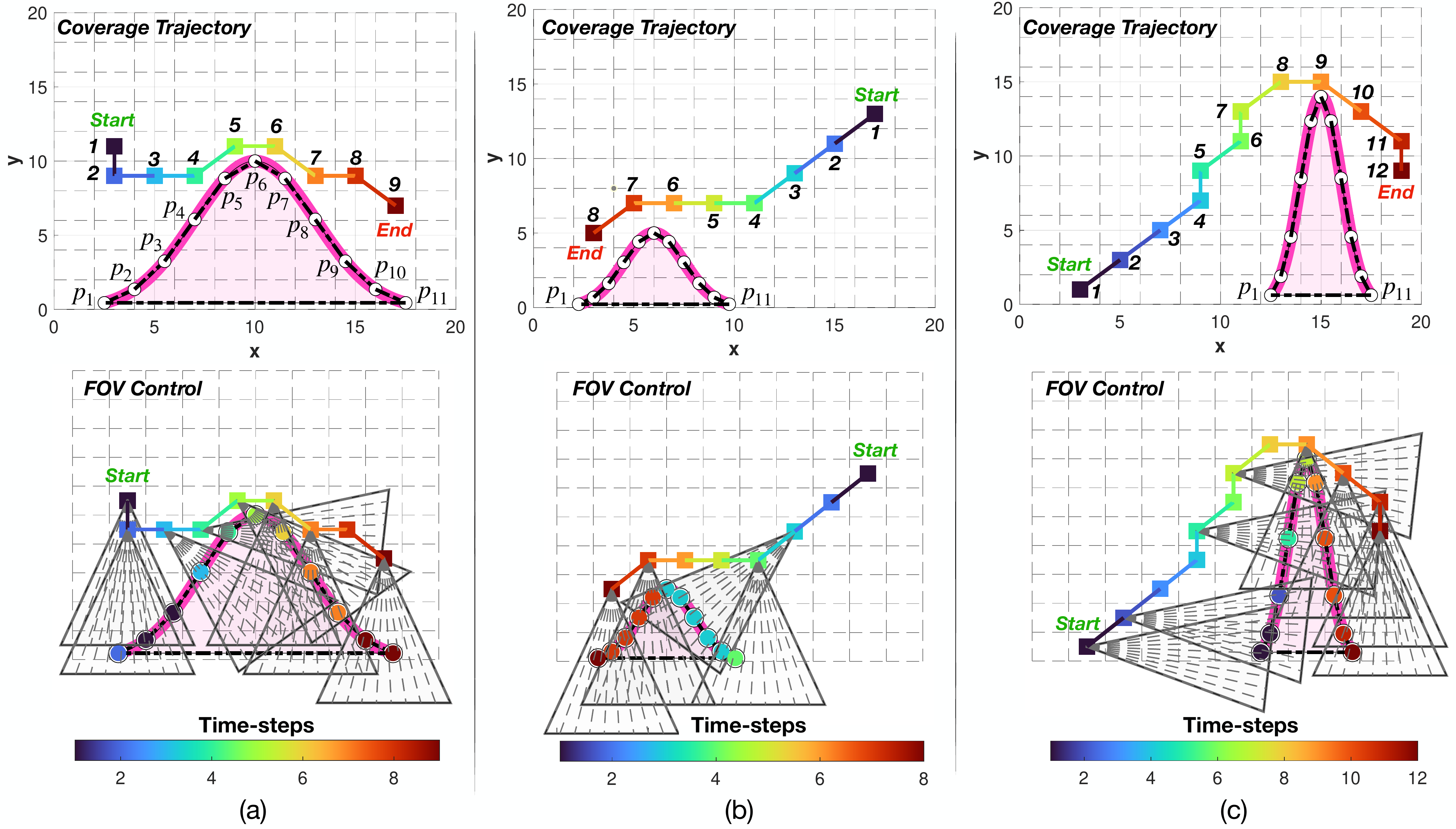}
	\caption{The figure illustrates 3 different scenarios (a), (b) and (c) for the problem of integrated ray-tracing and coverage planning control, where the agent follows the learned optimal control law $\pi_\star$.}	
	\label{fig:res3}
	\vspace{-6mm}
\end{figure*}

We begin our evaluation by investigating the learning performance of the proposed approach for a single and multiple objects of interest (i.e., different realizations from the same object class). Specifically, Fig. \ref{fig:res1}(a) shows the infinite horizon discounted cumulative reward per episode when the proposed system is trained over 5000 episodes for a single object of interest with parameters $(a,b,c) = (8,8,2)$. As the figure shows the episodic reward increases over time as the agent learns the optimal policy which  maximizes the state-action value function, in which case it reaches a plateau  as shown in the figure. Since the total cumulative reward obtained in each episode depends on the initial agent position, and because the agent in each episode is randomly initialized, the optimal control policy results in a cumulative reward which takes its values within a small range of values as shown in Fig. \ref{fig:res1}(a) after episode 2000. This is also evident in Fig. \ref{fig:res1}(b) which shows the total coverage time during training. As we can observe once the optimal policy is identified i.e., after 2000 episodes the agent requires on average 12 time-steps to finish the mission (depending on its initial state), whereas initially during the early episodes the mission is not finished not even within 100 time-steps.
Then Fig. \ref{fig:res1}(c) shows the learning curve also in terms of the discounted cumulative reward when the agent is trained over 100000 episodes for multiple realizations (i.e., parameter configurations) of the object of interest whose parameters $a, b$, and $c$ are uniformly sampled within the intervals $[1,18]$,  $[5,15]$ and $[1,4]$ respectively. In this scenario, for each episode a new realization of the object of interest is sampled and the agent is randomly initialized inside the surveillance area. An illustrative example of different object realizations sampled in each episode during training is shown in Fig. \ref{fig:res2}. Figure \ref{fig:res1}(c) indicates that the optimal policy can be identified and the state-action value function can be learned for multiple realizations of the object of interest. Figure \ref{fig:res3} shows in more detail the agent's coverage plan under the learned policy for 3 random realizations of the object of interest.
More specifically, Fig. \ref{fig:res3} depicts 3 different parameter configurations of the object of interest i.e., with the parameters $(a,b,c)$ set as $(10, 10, 2.5)$, $(5, 6, 1.4)$, and $(14, 15, 1)$ for the objects depiceted in Fig. \ref{fig:res3}(a), Fig. \ref{fig:res3}(b), and Fig. \ref{fig:res3}(c) respectively. In Fig. \ref{fig:res3} the first row illustrates the agent's kinematic states ($x_t$) under the learned control policy. The second row in addition to the agent's kinematic states, also shows the camera's FOV states $\mathcal{F}_t$ under the learned policy. The points $p_1,...,p_{11}$ on the object's boundary to be covered are shown as $\circ$. The agent's kinematic states $x_t$ (denoted with $\square$) are colored-coded and numbered based on their time-index $t$ as shown in the figure (i.e., from black to deep red, indicating start and end times respectively).

In addition, when the agent with kinematic state $x_t$ covers point $p$ at time $t$, we mark point $p$ with the same color as $x_t$, so that we can distiguish at which agent states each point is observed and covered. For instance in  Fig. \ref{fig:res3}(a), the agent at time-step $t=1$, with state $x_1$ (colored black) covers points $p_2$ and $p_3$, and therefore these points are colored black as well. At the next time-step the agent is at $x_2$ (colored blue) and observes point $p_1$ (note that $p_1$ resides inside the agent's camera FOV), thus $p_1$ is marked blue as shown. At $t=3$ the agent's state is colored cyan and so does point $p_4$ which is observed at that time, and so on and so forth. In order to make the figure easier to read, we only display the FOV configuration at time-steps for which a coverage event occurs e.g, in  Fig. \ref{fig:res3}(b) for instance the FOV configuration at time-steps $t=[1,2,5,6]$ is ommitted for clarity. 

Note that the control policy that the agent follows integrates ray-tracing. This is evident in Fig.\ref{fig:res3}(a) where at time-step $t=4$ the agent is at state $x_4$ (colored green) and points $p_5$ and $p_7$ reside inside the agent's triangular FOV. However, the agent only observes point $p_5$ as indicated by its color-coding i.e., point $p_5$ is colored green, whereas point $p_7$ is marked yellow and is not observed at $t=4$ since it is not visible from the agent's state $x_4$. Observe that the light-rays at $x_4$ are blocked by the object, and cannot be traced back to $p_7$. The point $p_7$ becomes visible and is covered when the agent moves at state $x_6$. This behavior can also be observed in Fig.\ref{fig:res3}(b) when the agent is at $x_3$, in Fig. \ref{fig:res3}(c) when the agent is at state $x_5$, etc.
Finally, note that the agent follows an optimal control policy which generalizes for multiple realizations of the object of interest, i.e., in all scenarios, all points are covered, the coverage time is minimized, and collisions between the agent and the object of interest are avoided. It is also worth noting that the optimal control policy has learned the optimal planning horizon length i.e., the coverage mission finishes in 9, 8 and 12 times-steps for the scenarios shown in  Fig. \ref{fig:res3}(a), Fig. \ref{fig:res3}(b), and Fig. \ref{fig:res3}(c) respectively, as opposed to the OCP formulation in which the optimal horizon length must be fined-tuned a-priori.

\section{Conclusion} \label{sec:conclusion}

In this work we have presented an integrated ray-tracing and coverage planning control approach which enables a mobile agent to jointly decide its kinematic and camera control inputs in such a way so that the total surface area of an object of interest is covered in the minimum amount of time. The proposed approach integrates ray-tracing into the coverage planning problem, in order to trace the propagation of light-rays back to their source thus ultimately identifying which parts of the scene are visible through the agent's camera. The problem is posed as a Markov decision process (MDP) and an optimal control law is obtained using reinforcement learning i.e., Q-learning. Future works include the investigation of this problem in 3D environments, its extension to multiple learning agents, and the adaptation of the proposed approach to continuous state-action spaces using deep reinforcement learning.

\section*{Acknowledgments}


This work is supported by the European Union's Horizon 2020 research and innovation programme under grant agreement No 739551 (KIOS CoE), by the European Union Civil Protection under grant agreement No 873240 (AIDERS), and from the Republic of Cyprus through the Directorate General for European Programmes, Coordination and Development.

\flushbottom
\balance

\bibliographystyle{IEEEtran}
\bibliography{IEEEabrv,main} 

\end{document}